\pdfoutput=1

\documentclass[11pt]{article}

\usepackage[final]{acl}
\usepackage{amsmath}
\usepackage{makecell}
\usepackage{changes}

\usepackage{amssymb}
\usepackage{float}
\usepackage{booktabs}  %
\usepackage{caption}   %

\usepackage{times}
\usepackage{latexsym}

\usepackage[T1]{fontenc}

\usepackage[utf8]{inputenc}

\usepackage{microtype}

\usepackage{inconsolata}

\usepackage{graphicx}

\title{AutoPsyC: Automatic Recognition of Psychodynamic Conflicts from Semi-structured Interviews with Large Language Models}

\author{Sayed Muddashir Hossain\textsuperscript{1}, Simon Ostermann\textsuperscript{1}, Patrick Gebhard\textsuperscript{1},\\
\textbf{Cord Benecke}\textsuperscript{2}, \textbf{Josef van Genabith}\textsuperscript{1}, \textbf{Philipp Müller}\textsuperscript{1}\\[0.5em]
\textsuperscript{1}DFKI, Saarbrücken, Germany\\
\textsuperscript{2}University of Kassel, Kassel, Germany\\[0.5em]
\texttt{\{sayed\_muddashir.hossain,simon.ostermann,patrick.gebhard\}@dfki.de}\\
\texttt{\{josef.van\_genabith,philipp.mueller\}@dfki.de}}

\begin{document}
\maketitle
\begin{abstract}
Psychodynamic conflicts are persistent, often unconscious themes that shape a person's behaviour and experiences.
Accurate diagnosis of psychodynamic conflicts is crucial for effective patient treatment and is commonly done via long, manually scored semi-structured interviews.
Existing automated solutions for psychiatric diagnosis tend to focus on the recognition of broad disorder categories such as depression, and it is unclear to what extent psychodynamic conflicts which even the patient themselves may not have conscious access to could be automatically recognised from conversation.
In this paper, we propose AutoPsyC, the first method for recognising the presence and significance of psychodynamic conflicts from full-length Operationalized Psychodynamic Diagnostics (OPD) interviews using Large Language Models (LLMs). 
Our approach combines recent advances in parameter-efficient fine-tuning and Retrieval-Augmented Generation (RAG) with a summarisation strategy to effectively process entire 90 minute long conversations.
In evaluations on a dataset of 141 diagnostic interviews we show that AutoPsyC consistently outperforms all baselines and ablation conditions on the recognition of four highly relevant psychodynamic conflicts.
\end{abstract}

\section{Introduction}

Accurate and detailed analysis of clinical interviews is essential for effective psychodynamic diagnostics. In particular, Operationalized Psychodynamic Diagnostics (OPD) interviews~\cite{force2008operationalized} serve as a cornerstone in psychodynamic assessment. 
A key aspect of OPD is the assessment of the patient's life-determining, often unconscious inner conflicts, such as conflicts relating to Dominance or Submissiveness, or to Self-value/esteem.
Automated analysis of psychodynamic conflicts from clinical interviews has the potential to support clinicians, reduce manual work, enhance objectivity, and may even lay the groundwork for the delivery of diagnostic interviews by artificial agents.
However, due to their long duration, low level of standardisation, and richness of information, semi-structured interviews pose unique challenges~\cite{adams2010joys, magaldi2020semi}. 
Prior natural language processing (NLP) work has often focused on short interview excerpts and broad diagnostic categories~\cite{low2020automated, milintsevich2023towards}.
To the best of our knowledge, no previous work has addressed the recognition of fine-grained psychodynamic concepts from long semi-structured diagnostic interviews.

In this work, we introduce a novel approach for recognising the presence and significance of psychodynamic conflicts as classified in the OPD from full-length interviews.
Our method combines recent advancements in parameter-efficient fine-tuning~\cite{hu2021lora} and Retrieval-Augmented Generation (RAG)~\cite{lewis2020retrieval} with a summarisation approach %
in order to process and classify long (90 min) semi-structured psychodynamic diagnostic interviews. 
In particular, we make use of a RAG framework to let the LLM access full-length interviews.
To allow the model to effectively reason about the interview to be scored, we additionally prompt it with a summary of the interview.
The classification is performed by an ensemble of four models, each of which was fine-tuned to analyse a specific temporal portion of an interview.
To evaluate our approach, we make use of a dataset of 141 OPD interview recordings~\cite{Bock2016}.
Our approach consistently improves over baselines and ablation conditions. 
It is able to reach weighted F1 scores of 0.78 and 0.81 for the conflicts \textit{Self-dependency and Dependency on Others}, and \textit{Dominance or Submissiveness}.
For the more challenging conflicts \textit{Self-sufficiency} and \textit{Self-value/esteem}, it is able to reach 0.59 and 0.58 F1, respectively.

Our specific contributions are threefold:

\begin{enumerate}
    \item We present AutoPsyC\footnote{Code available at \url{https://git.opendfki.de/philipp.mueller/autopsyc}}, the first LLM-based method for the recognition of presence and severity of psychodynamic conflicts from full-length OPD interviews, thereby bridging the fields of psychodynamic diagnostics and advanced NLP.%
    \item We evaluate AutoPsyC on a dataset of 141 90 minute long OPD interviews, showing that AutoPsyC consistently outperforms all baselines and in-depth ablation comparisons.
    \item We demonstrate that information contained in the middle sections of interviews is particularly informative for classifier training.
\end{enumerate}

\section{Related Work}

\subsection{Diagnostic Interviews in Psychotherapy and Psychiatry}

Structured interviews, using standardized questions and scoring, improve psychiatric diagnosis reliability by reducing clinician bias. Tools like the Structured Clinical Interview for DSM-5 (SCID-5, \citet{First2015}) ensure DSM-aligned accuracy but require extensive training, while the Mini-International Neuropsychiatric Interview (MINI, \citet{Sheehan1998}) offers efficient screening at the cost of some diagnostic precision. 
The Structured Interview of Personality Organization (STIPO) is a structured interview designed to assess personality functioning based on Kernberg’s object relations theory \cite{Clarkin2007}.

Unstructured interviews emphasize patient narratives and clinical intuition, enabling the exploration of unique experiences \cite{Nordgaard2013}. While fostering rapport and uncovering insights, their lack of standardization introduces variability and reduces reliability \cite{shea2016psychiatric, corbin2003unstructured, o2007unstructured, widiger2008clinical,Fava2024, Lenouvel2022}. In this context, the PDM-2 provides a multidimensional diagnostic framework that emphasizes psychological functioning and personality organization over categorical symptom-based diagnosis \cite{Lingiardi2015}. Likewise, Malan’s triangles offer a conceptual model for understanding intrapsychic conflict and resistance, rather than a formalized interview procedure \cite{Malan1979}.

Semi-structured interviews blend the structure of standardized questions with the flexibility to address emergent themes \cite{Fava2024, Lenouvel2022, adams2010joys, brinkmann2014unstructured, magaldi2020semi, adeoye2021research}. They have been shown to be particularly useful in complex cases like major depressive disorder \cite{Dupuy2020}. One example for a sem-structured format is the Core Conflictual Relationship Theme (CCRT) method to identify recurring interpersonal themes~\cite{Luborsky1998}.
Operationalized Psychodynamic Diagnosis (OPD) uses semi-structured methods rooted in psychodynamic theory to assess self-experience, interpersonal relationships, and unconscious conflicts \cite{force2008operationalized, cierpka2007operationalized}. 
Unlike symptom-focused tools, OPD provides in-depth insights into personality organization and internal dynamics, aiding personalized therapeutic interventions. 
Research has demonstrated the clinical relevance of OPD within therapeutic settings \cite{cierpka2007operationalized, Benecke2024, OPD2001, OPD2008}.
Despite their importance, the automatic analysis of semi-structured interviews remains under-explored.
In particular no previous work has attempted to automatically score OPD interviews.

\subsection{Large Language Models for Psychiatric Diagnosis}

The integration of NLP and machine learning in different aspects of mental health is a rapidly growing field of research~\cite{le2021machine,lindsay2021dissociating,hossain2024m3tcm}.
One particular focus of attention is the automated diagnosis of conditions such as depression or schizophrenia by analysing text, speech, and facial expressions \cite{barzilay2019predicting,low2020automated, kishimoto2022understanding, oh2024development, milintsevich2023towards,ettore2023digital}.
Tools like \textit{Diagnostisches Expertensystem für psychische Störungen }(DIA-X-5) are being tested for consistency \cite{hoyer2020test}, while patient involvement is emphasized to ensure ethical use \cite{brederoo2021implementation}.

Recent work has shown that LLMs can be utilised to analyse complex human affect expression in conversation~\cite{broekens2023fine,muller2024recognizing}, making them a promising candidate for applications in psychiatric disorders.
Indeed, LLMs are increasingly applied in psychiatry, identifying linguistic markers of disorders from social media posts and clinical transcripts \cite{farruque2024depression, lan2024depression, zhang2024llms}. These models also assist in parsing unstructured EHR notes for early diagnosis \cite{zhang2024llms}. %

However, current automatic methods mainly detect broad disorder categories without capturing 
the more detailed and often unconscious factors explored in psychodynamic assessments.
In particular, we are not aware of any approach to automatically recognise psychodynamic conflicts from clinical interviews.

\subsection{Integrating Domain Knowledge in Large Language Models}

The integration of domain-specific knowledge into LLMs, particularly through techniques like Retrieval-Augmented Generation (RAG), is transforming psychiatric applications by enabling models to access and apply current, relevant external information in real time \cite{lewis2020retrieval}. %
Building on RAG, RAFT (Retrieval-Augmented Facilitation for Text) further optimizes domain-specific knowledge integration by prioritizing the most relevant medical literature and clinical guidelines \cite{zhang2024raft}. 
Other methods, such as parameter-efficient domain knowledge integration and the use of human-annotated features, also improve LLM performance in biomedical contexts \cite{ke2021parameter, kim2024towards}. 
In our work, we utilise RAG techniques to present the first LLM-based system able to recognise psychodynamic conflicts expressed in full-length OPD interviews.

\begin{figure}[t!]
    \centering
    \includegraphics[width=\linewidth]{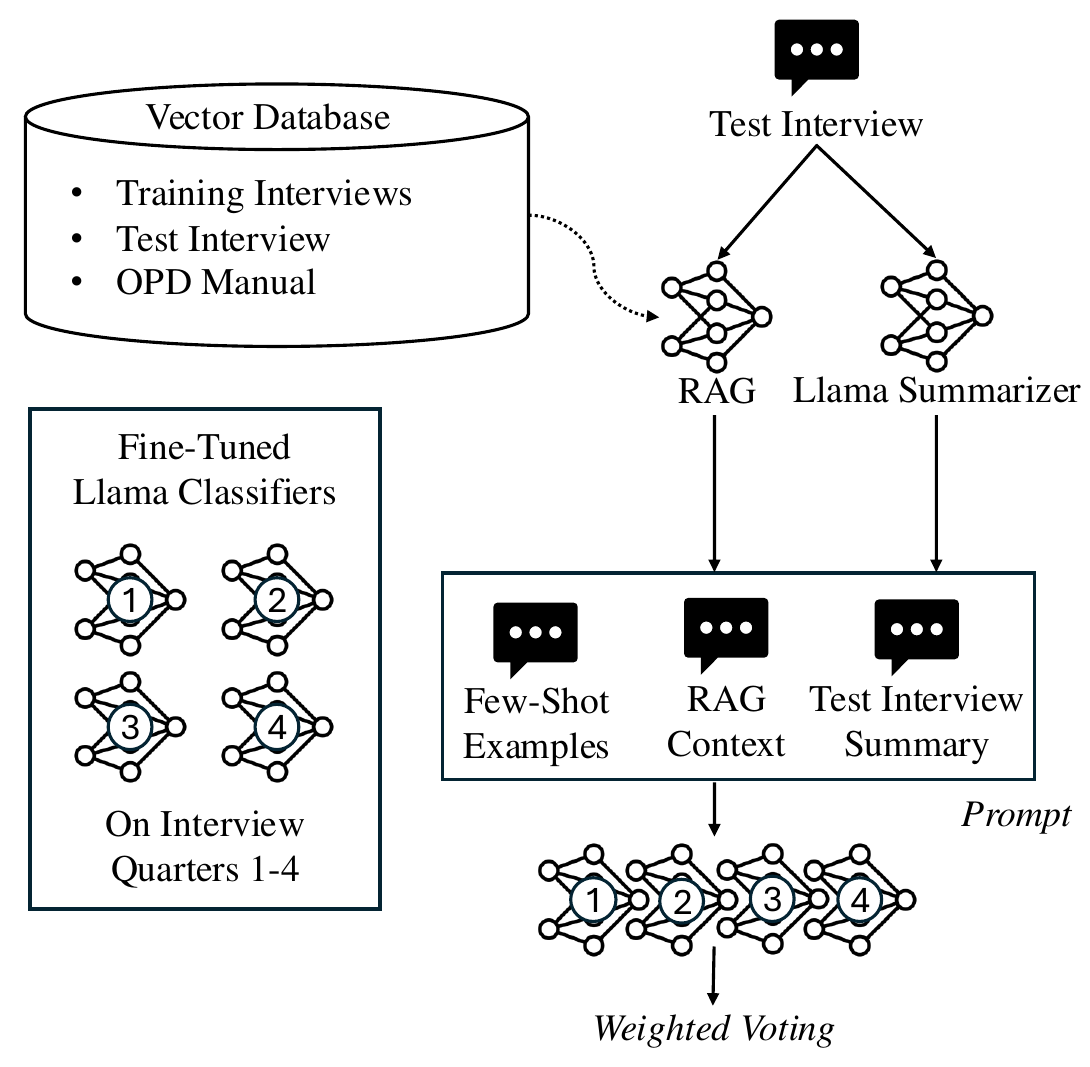}
    \caption{Overview over AutoPsyC.}
    \label{method}
\end{figure}

\section{Method}
A schematic overview of our method is shown in \autoref{method}.
The classification of a test interview consists of three steps.
In the first step, we employ LLaMA 3.1 (8B)~\cite{grattafiori2024llama3herdmodels} to generate a summary of the interview. 
In the second step, we use this summary to build a prompt for the final classification, which is performed by an ensemble of LLaMa 3.1 (8B) models that are fine-tuned to different interview segments.
The prompts for these specialised models contain 5 interview summaries from the training set, including ground truth (few-shot examples).
Via a RAG framework~\cite{lewis2020retrieval}, each specialised model also has access to relevant sections of the OPD manual, as well as to the full test interview and all full interviews from the training set.
In the third step, we combine the individual classifications obtained from the specialised models using a weighted voting scheme driven by a multinomial logistic regression.

\subsection{Summarization Method}
To obtain a focused representation of each interview, we first generate a summary using a LLaMA 3.1 8b model \cite{grattafiori2024llama3herdmodels}. The summarization prompt includes an example summary excerpted from the OPD Manual, instructing the model to adhere to a consistent style that reflects the diagnostic criteria. In this way, the generated summaries capture the diagnostically relevant information while filtering out extraneous details.

\subsection{Training Data Integration and RAG Setup}
In addition to the test interview summary, the classification prompt also includes few-shot examples in the form of summaries of interviews from the training set with associated ground truth labels.
We include one interview summary per ground truth class.
To further ground the classification in a domain-specific context, we employ a Retrieval-Augmented Generation (RAG) framework. 
In particular, we upload the following information into the RAG vector database.
\begin{enumerate}
    \item {Training Interviews:} We upload the full transcripts of all interviews from the training set without ground truth into the RAG knowledge base, pointing the model to them for retrieval. Adding ground truth information did not lead to improvements in preliminary experiments.
    \item {Test Interview:} We also upload the full transcript of the test interview.
    \item {OPD Manual:} The OPD-2 manual \cite{force2008operationalized} is organized into chapters corresponding to its axes, providing detailed descriptions and examples of OPD tasks and classifications. For our purposes, we included excerpts from the chapter on conflicts (Axis III) and the introductory section where the axes are defined and explained. Preliminary tests showed that incorporating the entire OPD-2 manual into the retrieval-augmented generation (RAG) system did not improve model performance compared to using only the relevant excerpts.
\end{enumerate}

This integration ensures that the classification model benefits from exemplars of each diagnostic class and explicit domain knowledge.
The model is also able to access detailed information present in the full interview transcripts in case the summaries are inconclusive.

\subsection{Classification Stage: Interview Segmentation and Finetuning}
Given that OPD interviews are semi-structured—with diagnostic cues distributed throughout—we split each interview (or its summary) into \( k \) segments, where in our implementation \( k = 4 \) (each segment being roughly 5,000 words).
In this way, each segments represents one quarter of the interview.
For each segment, we fine-tune a separate Llama 3.1 (8B) model using LoRA \cite{hu2021lora}, which allows for parameter-efficient adaptation. During training, we provide the model with a prompt including the segment summary, the RAG-augmented context (i.e., training interview summaries and manual excerpts), and an example for each of the five classes. The model is trained to output a probability distribution over the diagnostic classes. This process allows each fine-tuned model to capture the specific semantic and contextual nuances present in its corresponding interview segment.

\subsection{Result Aggregation}
After obtaining classification probabilities from each of the four fine-tuned models, we combine their outputs using a weighted voting scheme. Specifically, we train a multinomial logistic regression model that assigns a weight \( w_i \) to the prediction \( p_i(c) \) of the \( i \)-th segment for class \( c \). The final predicted class \( \hat{y} \) is computed as:
\begin{align}
\hat{y} = \arg\max_c \sum_{i=1}^k w_i\, p_i(c)
\end{align}

where \( k = 4 \) in our implementation.

\section{Dataset}

\begin{figure*}[ht!]
\centering
\includegraphics[width=1.0\linewidth]{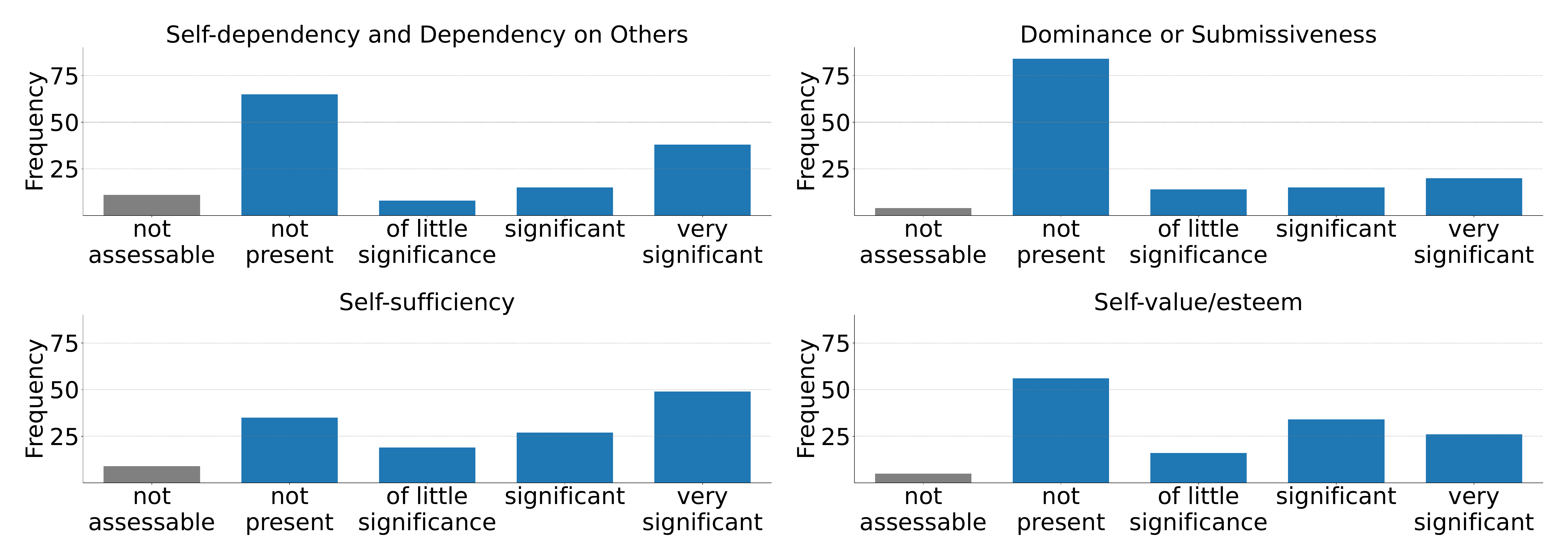}
\caption{Class distributions for the psychodynamic conflicts investigated in this study.}
\label{fig:output_classes}
\end{figure*}

The Kassel dataset~\cite{Bock2016}, utilized in this study, comprises 141 participants recorded during Operationalized Psychodynamic Diagnostics (OPD) interviews. %

\subsection{Participants}
The dataset includes both male (n = 21) and female (n = 120) participants, aged between 18 and 57 years.
Among them, 64 were inpatients diagnosed with at least one DSM-IV~\cite{APA2000} disorder, while 20 were healthy controls. The remaining participants had diverse diagnostic categories, including somatoform disorders (n = 22), borderline personality disorder (n = 19), depression (n = 18), and eating disorders such as anorexia (n = 14) and bulimia (n = 14). Anxiety disorders were observed in 13 participants. The inclusion criteria required informed consent, age above 18 years, and the absence of acute psychosis or schizophrenia.

\subsection{Data Collection}
Each participant underwent a clinical interview based on the framework of Operationalized Psychodynamic Diagnostics~\cite{force2008operationalized}. The interviews were carried out by a team of two male and two female interviewers~\cite{Bock2016}, all of whom were certified and trained in OPD application. The sessions, with an average duration of approximately 90 minutes, were recorded using split-screen technology to capture both the participants and interviewers. Both the interviewer and the interviewee were equipped with microphones to ensure clear audio capture. The audio from each session was transcribed verbatim into text by research staff or trained transcribers~\cite{Bock2016, vierl2023psychodynamic}. 
These recordings provided the foundational data for subsequent analyses of behavioural and contextual elements.

\subsection{Clinical Ground Truth}  

The dataset includes scores for Axes I-V of the OPD system.
In the scope of this paper, we focus on Axis III, which captures the patient's life-determining (un)conscious inner conflicts.
In particular, these are:
\emph{Conflicts related to self-dependency and dependency on others}, 
\emph{Conflicts associated with dominance or submissiveness}, 
\emph{Conflicts revolving around self-sufficiency}, 
\emph{Self-value and self-esteem conflicts}, 
\emph{Oedipal conflicts}, 
\emph{Identity-related conflicts}.
Each of these conflicts is rated with a five-class classification scheme.
The classes are \textit{Not assessable, Not present, Of little significance, Significant, Very significant}. Their detailed description can be found in~\ref{appendix:classes}.

We decided to omit the conflicts \emph{Oedipal Conflict} and \emph{Conflicts Related to Identity} from further analysis, as these conflicts were diagnosed in only a few instances, making a robust evaluation of predictions infeasible. 
For instance, in the case of \emph{Oedipal Conflict}, 120 out of 141 instances were labeled as \emph{not present} (see Appendix for further details). 
\autoref{fig:output_classes} illustrates the class distribution of the remaining conflicts.
While \textit{not present} is the most prevalent class for every conflict except Self-sufficiency, in all cases a significant portion of participants exist for whom the respective conflict is at least present with little significance.
In the following we provide a concise explanation of the four conflicts included in our analysis.

\textit{Conflicts related to self-dependency and dependency on others} refer to the tension between striving for autonomy and seeking support from others, often leading to struggles between independence and fear of isolation. \textit{Conflicts associated with dominance or submissiveness} involve power dynamics in interpersonal relationships, where individuals may oscillate between asserting control and yielding to authority, potentially resulting in power struggles or passivity. \textit{Conflicts revolving around self-sufficiency} pertain to the balance between the need for care and the desire for independence, with individuals experiencing inner turmoil when their reliance on others contradicts their self-reliance. \textit{Self-value and self-esteem conflicts} center on an individual's sense of worth, encompassing struggles with feelings of inadequacy or inferiority, often manifesting in compensatory behaviours aimed at reinforcing self-image. The conflicts were classified based on the overall interview. 
When binarizing the conflicts by treating \textit{not assessable} and \textit{not present} as “No,” and all other categories as “Yes,” only three cases did not present with any of the conflicts.

\section{Experiments}

\begin{table*}[ht]
\small
\resizebox{\textwidth}{!}{
\begin{tabular}{l|cccc}
 & \makecell{Self-dep. \\ \& \\ others-dep.} 
 & \makecell{Dom. \\ or \\ sub.} 
 & Self-suff. 
 & \makecell{Self-val. \\ \& \\ self-est.} \\ 
 \toprule
 \textit{Naive Baselines} \\
\ \ \ \ Demographic 
 & 0.31 ($\pm$0.01) 
 & 0.46 ($\pm$0.02) 
 & 0.31 ($\pm$0.03) 
 & 0.26 ($\pm$0.01) \\
\ \ \ \ Random 
 & 0.30 ($\pm$0.00) 
 & 0.33 ($\pm$0.00) 
 & 0.20 ($\pm$0.00) 
 & 0.23 ($\pm$0.00) \\
\midrule
\textit{No Training Data in VDB, No Fine-tuning} \\
\ \ \ \ w/o Manual 
 & 0.51 ($\pm$0.01) 
 & 0.63 ($\pm$0.01) 
 & 0.39 ($\pm$0.02) 
 & 0.43 ($\pm$0.03) \\
\ \ \ \ w/o Test Interv. in VDB 
 & 0.46 ($\pm$0.01) 
 & 0.60 ($\pm$0.01) 
 & 0.46 ($\pm$0.02) 
 & 0.48 ($\pm$0.01) \\
\ \ \ \ w/o Test Interv. Summary  
 & 0.53 ($\pm$0.01) 
 & 0.64 ($\pm$0.01) 
 & 0.42 ($\pm$0.02) 
 & 0.46 ($\pm$0.01) \\
 \ \ \ \ w/o Few-shot Examples  
 & 0.65 ($\pm$0.01) 
 & 0.68 ($\pm$0.01) 
 & 0.53 ($\pm$0.03) 
 & 0.48 ($\pm$0.02) \\
 \ \ \ \ AutoPsyC%
 & 0.68 ($\pm$0.01) 
 & 0.70 ($\pm$0.02) 
 & 0.55 ($\pm$0.01) 
 & 0.48 ($\pm$0.02) \\
\midrule
\textit{Training Data in VDB (Unlabelled), No Fine-tuning} \\
\ \ \ \ w/o Manual 
 & 0.48 ($\pm$0.03) 
 & 0.61 ($\pm$0.02) 
 & 0.43 ($\pm$0.02) 
 & 0.49 ($\pm$0.02) \\
\ \ \ \ w/o Test Interv. in VDB 
 & 0.50 ($\pm$0.01) 
 & 0.60 ($\pm$0.01) 
 & 0.45 ($\pm$0.01) 
 & 0.50 ($\pm$0.01) \\
\ \ \ \ w/o Test Interv. Summary 
 & 0.62 ($\pm$0.01) 
 & 0.69 ($\pm$0.02) 
 & 0.47 ($\pm$0.01) 
 & 0.52 ($\pm$0.02) \\
 \ \ \ \ w/o Few-shot Examples 
 & 0.68 ($\pm$0.02) 
 & 0.73 ($\pm$0.01) 
 & 0.57 ($\pm$0.01) 
 & 0.50 ($\pm$0.04) \\
 \ \ \ \ AutoPsyC%
 & 0.70 ($\pm$0.01) 
 & 0.74 ($\pm$0.02) 
 & 0.58 ($\pm$0.01) 
 & 0.50 ($\pm$0.02) \\
\midrule
\textit{Training Data in VDB (Unlabelled), Fine-tuning} \\
 \ \ \ \ w/o Test Interv. Summary \& Manual \& Train Interv. in VDB
 & 0.65 ($\pm$0.04) 
 & 0.68 ($\pm$0.01) 
 & 0.49 ($\pm$0.02) 
 & 0.47 ($\pm$0.02) \\
 \ \ \ \ w/o Test Interv. Summary \& Manual
  & 0.69 ($\pm$0.02) 
 & 0.72 ($\pm$0.03) 
 & 0.50 ($\pm$0.01) 
 & 0.49 ($\pm$0.01) \\
\ \ \ \ w/o Test Interv. Summary \& Weighted Voting 
 & 0.73 ($\pm$0.01) 
 & 0.75 ($\pm$0.02) 
 & 0.56 ($\pm$0.01) 
 & 0.53 ($\pm$0.02) \\
\ \ \ \ w/o Test Interv. Summary \& Ensemble 
 & 0.72 ($\pm$0.02) 
 & 0.74 ($\pm$0.01) 
 & 0.55 ($\pm$0.02) 
 & 0.52 ($\pm$0.01) \\
\ \ \ \ w/o Manual \& Train Interv. in VDB
 & 0.68 ($\pm$0.02) 
 & 0.72 ($\pm$0.01) 
 & 0.51 ($\pm$0.02) 
 & 0.49 ($\pm$0.01) \\
\ \ \ \ w/o Weighted Voting 
 & 0.75 ($\pm$0.02) 
 & 0.78 ($\pm$0.01) 
 & 0.55 ($\pm$0.02) 
 & 0.57 ($\pm$0.01) \\
\ \ \ \ w/o Ensemble 
 & 0.71 ($\pm$0.02) 
 & 0.74 ($\pm$0.01) 
 & 0.56 ($\pm$0.02) 
 & 0.55 ($\pm$0.01) \\
\ \ \ \ w/o Train Interv. in VDB
 & 0.74 ($\pm$0.02) 
 & 0.77 ($\pm$0.01) 
 & 0.56 ($\pm$0.02) 
 & 0.55 ($\pm$0.01) \\
 \ \ \ \ w/o Manual 
 & 0.72 ($\pm$0.01) 
 & 0.74 ($\pm$0.02) 
 & 0.54 ($\pm$0.01) 
 & 0.52 ($\pm$0.02) \\
 \ \ \ \ w/o Test Interv. in VDB 
 & 0.70 ($\pm$0.02) 
 & 0.73 ($\pm$0.01) 
 & 0.53 ($\pm$0.02) 
 & 0.50 ($\pm$0.01) \\
 \ \ \ \ w/o Test Interv. Summary 
 & 0.75 ($\pm$0.01) 
 & 0.80 ($\pm$0.02) 
 & 0.57 ($\pm$0.01) 
 & 0.57 ($\pm$0.02) \\
 \ \ \ \ w/o Few-shot Examples
 & 0.73 ($\pm$0.01) 
 & 0.74 ($\pm$0.02) 
 & 0.55 ($\pm$0.01) 
 & 0.51 ($\pm$0.02) \\
\ \ \ \ \textbf{AutoPsyC} 
 & \textbf{0.78 ($\pm$0.02)} 
 & \textbf{0.81 ($\pm$0.01)} 
 & \textbf{0.59 ($\pm$0.02)} 
 & \textbf{0.58 ($\pm$0.01)} \\
 \bottomrule
\end{tabular}}
\caption{Average Weighted F1-Scores with 95\% Confidence Intervals. 
}
\centering
\label{tab:results}
\end{table*}

In this section, we describe the evaluation protocol and baselines.

\subsection{Evaluation Protocol}

The dataset was partitioned into five fixed folds using stratified 5-fold cross-validation to maintain the proportional representation of key demographic and diagnostic variables. This was achieved by utilizing the \texttt{StratifiedKFold} module from the \texttt{scikit-learn} library~\cite{scikit-learn}, with stratification based on \textit{Gender} and \textit{Diagnosis}. The stratification process guarantees a fair distribution of these attributes across all folds. This consistency was preserved across all experiments involving the 5-fold cross-validation framework.

To evaluate all models and baselines, we make use of the weighted F1 score. The weighted F1 score accounts for class imbalance by weighting classes proportionally to their prevalence, ensuring robust evaluation of both frequent and rare diagnostic categories. It balances precision (avoiding overpathologizing) and recall (preventing missed conflicts), aligning with clinical priorities. 
To robustly estimate performance, we repeated all experiments several times and report the average weighted F1 score across all runs.
In the case of experiments involving LLMs, we average across 100 runs, and in the case of the computationally less expensive baseline experiments, we average across 1000 runs.
In addition to the averages, we also report their 95\% confidence interval.

\subsection{Baselines}

We implement two simple baselines: a \textit{Demographic Baseline} and a \textit{Random Baseline}. %

\paragraph{Demographic Baseline.}
This baseline employs a neural network classifier using demographic attributes such as gender, clincial diagnosis group, and binned age as input features. Numerical features were normalized by subtracting the mean and dividing by the standard deviation, while categorical features were converted into numerical representations. The neural network, implemented in PyTorch~\cite{paszke2019pytorchimperativestylehighperformance}, consists of three fully connected layers with ReLU activations. It was trained for 30 epochs using cross-entropy loss and the Adam optimizer.

\paragraph{Random Baseline.} %
The random baseline leverages the \texttt{DummyClassifier} module from \texttt{scikit-learn}, configured with the \texttt{stratified} strategy. This classifier generates predictions by randomly assigning labels based on the class distribution of the training set. This random classifier serves as a naive baseline, highlighting the minimum expected performance for the classification task.

\section{Results and Discussion}

\subsection{Overview}
Table~\ref{tab:results} summarises our weighted F\textsubscript{1} scores across the four psychodynamic conflicts. To more easily navigate the table, we partition the different ablation conditions into three cases, based on whether unlabelled training data is encorporated in the RAG framework and based on whether fine-tuning is performed with labelled training data.
Ablations are always named relative to the partition their are in. 
For example, the ablation \textit{w/o Manual} in the partition \textit{No Training Data in VDB, No Fine-tuning} describes an ablation condition without training data integration into the RAG framework, without fine-tuning, and without integration of the OPD Manual in the RAG framework.

\begin{figure}[ht]
\centering
\includegraphics[width=\linewidth]{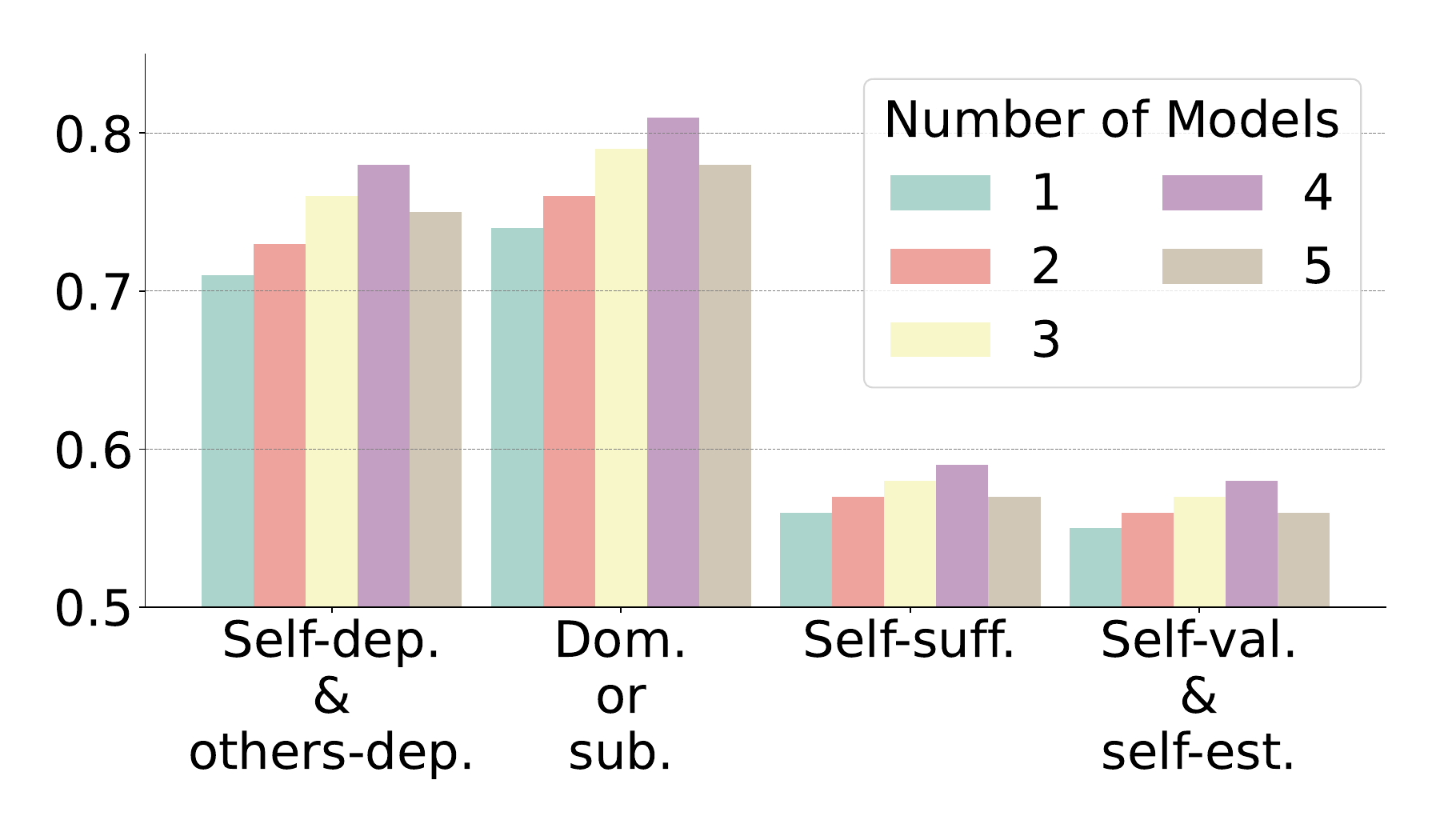}
\caption{Performance of Different Number of Models}
\label{fig:diff_model_performance}
\end{figure}

We observe that our full method (AutoPsyC), which combines our summarisation strategy with weighted voting across fine-tuned, temporally specialised models, as well as domain knowledge integration into the RAG framework, consistently outperforms all baselines and ablation conditions.
As illustrated in Figure~\ref{fig:model_performance}, models fine-tuned on the middle segments of the interviews consistently outperform those focusing on earlier or later sections. 
Moreover, Figure~\ref{fig:diff_model_performance} indicates that deviating from four total models or partitioning the interviews into fewer or more than four segments leads to a noticeable drop in overall performance.
Finally, we present an analysis of gender fairness of our model.
Overall low values of Conditional Demographic Disparity (CDD) indicate no major gender-related biases (\autoref{tab:cdd-values}).

\subsection{Which Model Configuration Works Best?}
Our experimental results demonstrate the effectiveness of combining our summarisation strategy with Retrieval-Augmented Generation (RAG), instruction tuning and section-wise model specialization for psychodynamic conflict classification in clinical interviews. 
As can be seen in (Table~\ref{tab:results}), AutoPsyC achieves superior performance across all four conflict categories, with weighted F1-scores ranging from 0.58 to 0.81. This represents a substantial improvement over both naive baselines (Demographic: 0.26–0.46; Dummy: 0.20–0.33) and non-instruction-tuned variants (0.50–0.74).

Our detailed ablation experiments indicates that AutoPsyC effectively integrates all available information.
We can observe a large decrease in performance when the test interview transcript is removed from the vector database (0.50-0.73 F1).
This indicates that our model indeed makes use of the full test interview transcript that is provided via the RAG framework to fill in information missing in the interview summary.
At the same time, we see that it does profit from the test interview summary, with losses of up to 0.04 F1 when the summary is removed.
We furthermore observe a clear loss in performance when the OPD manual is removed from the vector data base (0.52-0.74 F1), and a slightly lower loss in performance when the training set interviews are removed from the database (0.55-0.77 F1).
This indicates that even when using fine-tuned classification model, domain knowledge integration via the RAG setup is still helpful.
The weighted voting mechanism using multinomial logistic regression provides moderate but consistent performance gains (0.58–0.81 F1 vs. 0.55–0.78 for unweighted aggregation), suggesting that different interview sections contribute asymmetrically to conflict identification.

One general observation we can make is that fine-tuning leads to greater robustness w.r.t. other ablation conditions.
E.g. removing the OPD Manual from the vector database leads only to a moderate loss in performance when fine-tuned classification models are used (0.58-0.81 F1 vs. 0.52-0.74 F1).
In contrast, for the case of no fine-tuning, the losses are more dramatic (0.50-0.74 F1 vs. 0.43-0.61 F1).

\subsection{Which Interview Section is most useful?}
\begin{figure}[ht]
\centering
\includegraphics[width=1\linewidth]{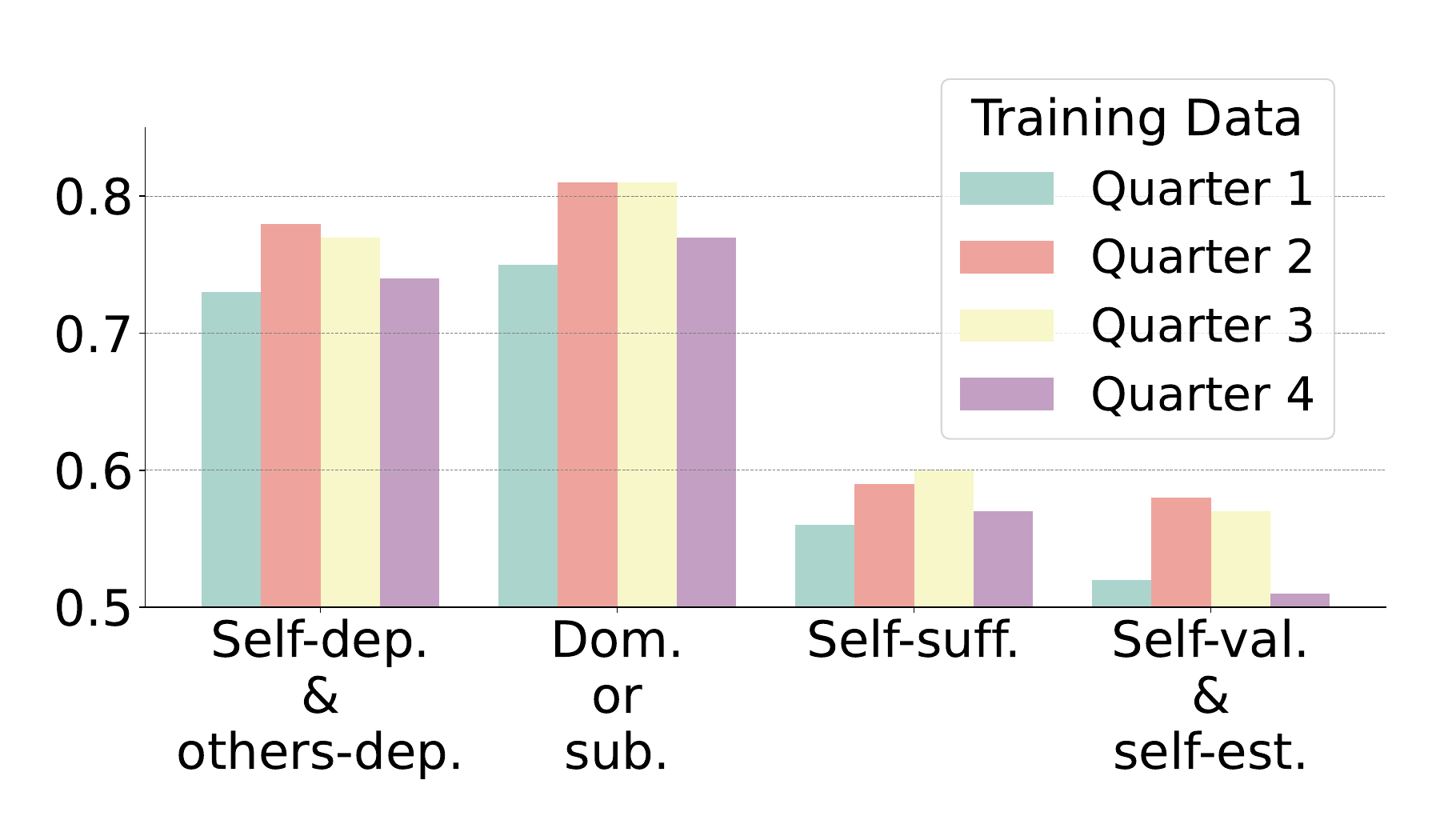}
\caption{Performance of the four models across all classes.}
\label{fig:model_performance}
\end{figure}

To further investigate which sections of the interviews are most informative fine-tuning classification models, we investigate the performance of our four pretrained models singled out across all conflicts (see \autoref{fig:model_performance}).
The results indicate that the models fine-tuned using the middle sections of the interviews outperform those tuned with other sections. %
After a careful examination of the interviews, we observed that, in the quarter 2 \& 3, the interviewees often share information more closely related to their condition and situation. 
An excerpt of the interview can be found in Appendix~\ref{appendix:transcript}.

\begin{table*}[ht]
    \centering
    \small
    \begin{tabular}{lcccc}
    \toprule
    & \textit{Self-dep.\ \& others-dep.} & \textit{Dom.\ or sub.} & \textit{Self-suff.} & \textit{Self-val.\ \& self-est.} \\
    \midrule
    not assessable         & 0.0031 & 0.0008 & 0.0053 & 0.0044 \\
    not present            & 0.0042 & 0.0023 & 0.0034 & 0.0014 \\
    of little significance & 0.0018 & 0.0011 & 0.0019 & 0.0020 \\
    significant           & 0.0025 & 0.0021 & 0.0054 & 0.0009 \\
    very significant      & 0.0029 & 0.0015 & 0.0010 & 0.0027 \\
    \bottomrule
    \end{tabular}
    \caption{One-vs.-rest CDD values for each class across four conflicts.}
    \label{tab:cdd-values}
\end{table*}

\subsection{Additional Experiments: How Fair is the Model?}

Fairness concerning gender is a critical issue in psychiatric diagnoses when using machine learning algorithms, as biases in training data or model predictions can lead to systematic disparities in diagnostic outcomes. 
For instance, a study by~\citet{Mosteiro2022Bias} found that gender played an unexpected role in predictions related to benzodiazepine administration, potentially biasing the model's decisions.

We evaluate fairness with respect to the gender attribute using Conditional Demographic Disparity (CDD)~\cite{wachter2020why}.
CDD quantifies the difference in expected outcomes across demographic groups, with values closer to zero indicating fairer conditions. 

\begin{align}
\text{CDD} = \mathbb{E}[\hat{y} \mid  \text{male}, y] - \mathbb{E}[\hat{y} \mid \text{female}, y]
\end{align}

Since we have more than two classes we compute CDD in one-vs.-rest fashion. We compute 
\begin{equation}
\text{CDD}_k = 
\mathbb{E}\bigl[\hat{p}^k \mid \text{male},\, y\bigr]
\;-\;
\mathbb{E}\bigl[\hat{p}^k \mid \text{female},\, y\bigr]
\end{equation}
for class \(k\). This formulation reduces the multi-class problem to a 
``one vs.\ rest'' scenario by focusing on a single predicted probability 
\(\hat{p}^k\). If class \(k\) is deemed the ``positive'' class, the 
binary-based fairness thresholds from~\cite{wachter2020bias} can be applied to \(\text{CDD}_k\).

Overall, the CDD values reported in Table~\ref{tab:cdd-values} are relatively low (all below 0.006), as indicated by prior studies~\cite{wachter2020why,koumeri2023compatibility,wachter2020bias}, suggesting minimal gender-based disparity across the different classes. \citet{wachter2020why} suggest that CDD values below 0.01 indicate minimal demographic disparity, while \citet{koumeri2023compatibility} and ~\citet{wachter2020bias} provide empirical evidence that values above 0.02-0.05 often indicate fairness concerns.
It is important to note that this fairness evaluation is not able to account for potential biases that are already present in the ground truth annotations on the dataset.

\subsection{Ethical Considerations and Impact}

The automation of psychodynamic diagnostics using NLP and machine learning presents both opportunities and ethical challenges. While enhancing objectivity, efficiency, and accessibility, its implementation requires careful ethical scrutiny to ensure responsible use in mental health care. Psychodynamic interviews contain sensitive data, necessitating strong anonymization and compliance with privacy regulations such as GDPR and HIPAA. Additionally, automated diagnostics may reflect biases present in training data, leading to disparities across demographic groups. Continuous bias auditing and fairness assessments are essential to mitigate these risks and ensure equitable model performance.

Automated tools should complement, not replace, human expertise. 
AutoPsyC could serve as a supplementary tool for therapists during psychodynamic interviews, acting as a "second-eye" to enhance clinical decision-making \cite{apa2025_ai_mentalhealth}.
Additionally, AutoPsyC could be integrated into social interactive agents, chat applications, and telepsychiatry platforms, providing complementary therapeutic support \cite{smith2019_telehealth_ai}. Furthermore, it could be utilized in psychological training tools to enhance the proficiency of conducting psychodynamic interviews \cite{apa2023_ai_training}.

Psychodynamic diagnostics involve complex interpretations that extend beyond text-based pattern recognition. Thus, model outputs must be interpretable, allowing clinicians to integrate them into their assessments. Future research should prioritize explainability and transparency in AI-driven diagnostics.
As AI applications in mental health expand, concerns arise regarding consent, misuse, and potential stigmatization in non-clinical settings. Interdisciplinary collaboration among clinicians, ethicists, and policymakers is needed to safeguard patient autonomy and well-being.

\section{Conclusion}
We present a novel framework for automated conflict classification in psychodynamic interviews, achieving clinically relevant performance through three key innovations: (1) domain-adapted instruction tuning using segmented interview data, (2) RAG-enhanced contextual understanding through OPD Manual and other interview integration, and (3) confidence-weighted aggregation of specialized section models.

These results suggest that LLMs can be effectively adapted for complex psychiatric coding tasks when combined with domain-specific knowledge retrieval and structured interview analysis. The demonstrated technique for identifying diagnostically salient interview segments (quarters 2 \& 3) offers methodological insights for computational psychiatry research. Future work should explore applications to other OPD Axes and integration with multimodal clinical data.

\section{Limitations}
While promising, our approach has several limitations. First, the dataset size (n=141 interviews) may limit generalizability, particularly for rare conflict subtypes. Second, the complex pipeline (RAG, summarization, 4 specialized models) incurs significant computational costs compared to monolithic models. Third, performance variation across conflict categories (0.58–0.81 F1) suggests task-specific optimization needs, particularly for \textit{Self-sufficiency} classification.

The reliance on manual OPD Manual examples for summarization introduces potential annotation bias, and the gender fairness analysis does not account for non-binary identities. Additionally, our stratified sampling based on diagnosis and gender may not fully capture all confounding demographic factors. %
We focus on a single summarization approach, as our primary goal is to establish a proof of concept for automated OPD scoring. While alternative summarization methods could be explored in future work, this choice allows us to maintain methodological consistency and provide a clear baseline for comparison.

We split the interview into four parts based on word counts, which does not fully account for the semi-structured nature of our interviews. Future work could focus on automatically detecting interview segments for the fine-tuning process.

Our fairness analysis showed minimal gender-based disparity in predictions.
There are however many other ways in which our model may be biased.
On the interview dataset we utilised, we did not have access to e.g. information on socioeconomic status or education.
Further variations are such as cultural background are not sufficiently covered by the dataset as it was recorded with German-speaking people in Europe.
This geographic and cultural constraint represents another key limitation of our study.
It remains unclear, whether our approach would also work in very different cultural contexts.

Future research should address these limitations through larger multicentre datasets\added{~\cite{konig2022multimodal}}, lightweight model architectures, and explicit modeling of clinician raters' variance. The current implementation also requires further validation for real-time clinical deployment, including robustness testing against speech recognition errors and patient dialect variations. 
A more detailed investigation of how AutoPsyC handles defensive processes~\cite{freud1936abwehrmechanismen} remains an area for future research.

\bibliography{custom}

\newpage
\appendix
\section{Appendix}
\label{sec:appendix}
\subsection{Excerpt from the Interview Transcript}
\label{appendix:transcript}
\subsection*{Beginning}
\begin{quote}
\textbf{Interviewer:} So, let us begin with the second part. I will ask you various questions from different areas. \\

\textbf{Interviewee:} Hmm. \\

\textbf{Interviewer:} About your present life, your past, relationships, work life, etc. Yes. I would like to start by asking you to describe what is currently the most burdensome for you. \\

\textbf{Interviewee:} At the moment? \\

\textbf{Interviewer:} Yes, it can be anything. \\

\textbf{Interviewee:} (Exhales) Ahm. (-) In general, I am actually doing quite well. However, I must say that things that have burdened me in the past, especially over the past three years, have now become less significant. What currently affects me the most is the situation at home. My parents are about to get divorced, and that has been the most difficult thing in my life so far. I must say, it has also been very stressful for me, but I am slowly managing it quite well. \\
\end{quote}

\subsection*{Middle}
\begin{quote}
\textbf{Interviewee:} The period was simply long. I was 20 when it all started, and I would say that only in the last few months have I truly felt lighter inside. For about a year, things have been steadily improving, but before that, I felt terrible. At home, it was a crisis. My mother was struggling—she barely ate, she was just existing. That made me very sad because I am someone who tries to keep everyone together. Given my age, I was able to grasp everything more clearly. I always spoke with everyone, tried to mediate, and made sure we somehow lived through it. But it was simply too much. (Claps hands on the table.) \\

I do not regret anything, or at least not much, except for the lingering aftereffects, which sometimes scare me. But otherwise, I would do it all over again. It just went too far. There were long periods where I barely met anyone or made any plans. If someone invited me out, I would always say no because I had to check on my mother to see if she was alone. \\

It was a responsibility that suddenly fell upon me. I would not say that it was forced upon me—I took it on willingly. That is simply the kind of person I am. If I see someone struggling, I cannot ignore it. I am very attached to my family. \\

There were times, for example, at Easter, when my mother just drove off. I could see in her eyes that she did not want to live anymore. She says the same thing even now. Back then, it was even stronger—she simply did not want to go on. She just got into the car and drove away. (Shocked and saddened.) It was simply terrible. \\

At first, I wanted to prevent the separation, of course. As a child, you never want your parents to separate. But later, it was just about minimizing the damage. I lost count of how many times I sat there listening, trying to mediate. I took on the role of always being there. But at some point, it was just too much. \\

I still managed to get through it, though sometimes I look back and wonder how I did it. I held up well, except for my university studies, where I had some setbacks. That was where the burden really showed. The emotional toll and the time commitment were simply too much. \\
\end{quote}

\subsection*{End}
\begin{quote}
\textbf{Interviewee:} I actually feel much better now. I have accepted everything as it is. I am a realistic person. I do not try to convince myself of things that do not exist. I walk through life with open eyes. I see what is happening around me. I know the divorce statistics. \\

\textbf{Interviewer:} But until recently, they did not matter to you. \\

\textbf{Interviewee:} No. \\

\textbf{Interviewer:} (Laughs.) \\

\textbf{Interviewee:} (Laughs as well.) Yes, because within those four walls, everything was fine. That was my foundation, my roots, where I came from. It was intact. \\

\textbf{Interviewer:} But now that has changed. \\

\textbf{Interviewee:} Yes. And I know that no matter how well things may seem to be going, there is always the risk that it could fall apart. That belief, that certainty I once had, is gone. I used to truly believe in lasting relationships. But now, if you ask me whether I think a relationship will last a lifetime, I no longer believe that. It is a sad realization. \\

\textbf{Interviewer:} It sounds as if a vision or a dream has been lost. \\

\textbf{Interviewee:} Yes, definitely. No doubt about it. \\
\end{quote}

\subsection{Example Prompt for \textit{Self-dependency and Dependency on Others}}
\label{appendix:example}
\textbf{Context:} Relationships and attachments are of existential importance in every person’s life. They span the opposing poles of striving for close relationships and symbiotic proximity (dependency) and striving for well-developed independence and clear distance (powerful individuation). Individuation and dependency are fundamental elements of human life and experience, present in all areas of life. A life-defining conflict arises when this fundamental bipolar tension turns into a conflictual polarization. An individuation-dependency conflict is present only if this constellation is of existential importance and formative for a person's life history: This conflict involves the activation of experiences that either seek or avoid closeness, not the shaping of relationships in terms of caregiving or avoiding caregiving. The theme of individuation-dependency deals with the question of being alone or the ability to be with others. In its pathological conflict version, it concerns the necessity of being alone or being with others as an existential requirement.

\textbf{Task:} Based on this context, classify the following interview excerpt regarding the theme of "autonomy-dependency" into one of the following categories: \emph{"not present"}, \emph{"not assessable"}, \emph{"of little significance"}, \emph{"significant"}, \emph{"very significant"}.
For tasks where the interviews were summarized prior to classification, the model was first instructed to generate a summary of the interview based on a provided example. This example was derived from the OPD Manual.

\subsection{Classification Classes}
\label{appendix:classes}
\begin{itemize}
    \item \textbf{Not assessable} – The category cannot be determined due to insufficient or ambiguous information. There may be a lack of relevant content, unclear statements, or methodological limitations preventing a reliable assessment.
    
    \item \textbf{Not present} – There is no indication of the characteristic or phenomenon being evaluated. The available information does not support its existence or relevance in the given context.
    
    \item \textbf{Of little significance} – The characteristic or phenomenon is present but plays only a minor role. It appears occasionally but does not have a substantial influence on behavior, emotions, or interactions.
    
    \item \textbf{Significant} – The characteristic or phenomenon is clearly identifiable and has a notable impact. It influences thoughts, emotions, or interactions and is relevant to the overall assessment.
    
    \item \textbf{Very significant} – The characteristic or phenomenon is a dominant feature. It strongly shapes experiences, interactions, or coping mechanisms and is central to the evaluation.
\end{itemize}

\subsection{Extra Plots}
\label{appendix:plots}
\begin{figure}[ht!]
\centering
\includegraphics[width=1\linewidth]{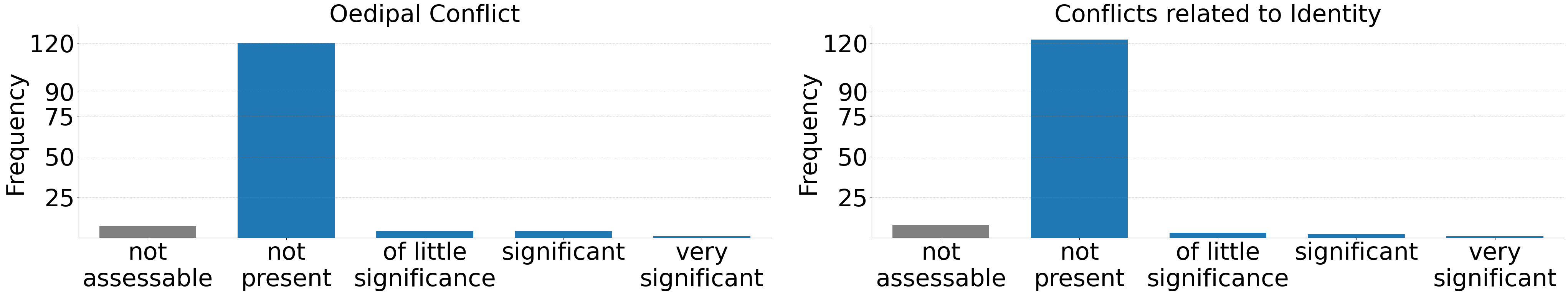}
\caption{Class distributions for Oedipal Conflict and Conflicts related to Identity}
\label{fig:not_output_classes}
\end{figure}

\end{document}